# Graph Attention Networks Unveil Determinants of Intra- and Inter-city Health Disparity

Chenyue Liu[1*], Chao Fan[2], Ali Mostafavi[3]

[1] Ph.D. Student, Urban Resilience.AI Lab, Zachry Department of Civil and Environmental Engineering, Texas A&M University, College Station, United States; e-mail: liuchenyue@tamu.edu

[2] Assistant Professor, Glenn Department of Civil Engineering, Clemson University, Clemson, South Carolina, United States; e-mail: cfan@g.clemson.edu

[3] Associate Professor, Urban Resilience.AI Lab Zachry Department of Civil and Environmental Engineering, Texas A&M University, College Station, United States; e-mail: amostafavi@civil.tamu.edu

**Declarations of interest:** none

**Acknowledgement**

This material is based in part upon work supported by Texas A&M University X-Grant 699. The authors also would like to acknowledge that Spectus and SafeGraph provided mobility and population activity data. Any opinions, findings, conclusions or recommendations expressed in this material are those of the authors and do not necessarily reflect the views of the Texas A&M University, Spectus and SafeGraph.

# Abstract

Understanding the determinants underlying variations in urban health status is important for informing urban design and planning, as well as public health policies. Multiple heterogeneous urban features could modulate the prevalence of diseases across different neighborhoods in cities and across different cities. This study examines heterogeneous features related to socio-demographics, population activity, mobility, and the built environment and their non-linear interactions to examine intra- and inter-city disparity in prevalence of four disease types: obesity, diabetes, cancer, and heart disease. Features related to population activity, mobility, and facility density are obtained from large-scale anonymized mobility data. These features are used in training and testing graph attention network (GAT) models to capture non-linear feature interactions as well as spatial interdependence among neighborhoods. We tested the models in five U.S. cities across the four disease types. The results show that the GAT model can predict the health status of people in neighborhoods based on the top five determinant features. The findings unveil that population activity and built-environment features along with socio-demographic features differentiate the health status of neighborhoods to such a great extent that a GAT model could predict the health status using these features with high accuracy. The results also show that the model trained on one city can predict health status in another city with high accuracy, allowing us to quantify the inter-city similarity and discrepancy in health status. The model and findings provide novel approaches and insights for urban designers, planners, and public health officials to better understand and improve health disparities in cities by considering the significant determinant features and their interactions.

# Introduction

Understanding the effects of urban neighborhood factors on residents' health is one critical factor for addressing health disparities in cities. In particular, understanding the extent to which urban features can explain the risk of disease within and across cities is an essential tool for integrated urban design that promotes public health. Previous literature has considered a wide range of determinants of health status, but the majority of existing studies consider only limited types of determinants. For example, (Galiatsatos et al., 2018) examined social-economic determinants to explain the relationship with health outcomes at the neighborhood level. (Subramanian & Kawachi, 2003) examine the association between state income inequality and poor self-rated health. This study compared physician use in Ontario and the midwestern and northeastern United States for persons of different socioeconomic status and health(Katz et al., 1996). (Mason et al., 2018) reported the connection between high densities of physical activity facilities and lower obesity for adults in mid-life. (Farmer & Ferraro, 2005) examine health disparities between white and black adults and whether the SES/health gradient differs across the two groups in the USA. (Michael et al., 2014) evaluate the effect of a neighborhood-changing intervention on changes in obesity among older women. (Tang et al., 2022) perform principal component regression (PCR) to assess the relationship between the built environment and both self-rated physical health and mental health.(Hasthanasombat & Mascolo, 2019) propose an approach to link the effects of neighborhood services over citizen health using a technique that attempts to highlight the cause-effect aspects of these relationships. (Wang et al., 2017) link four models to evaluate the effects of both international exports and interprovincial trade on PM pollution and public health across China. Recently, a number of studies have used urban human mobility data as determinants of public health status. (Bauer & Lukowicz, 2012) described a machine learning-based system that estimated well-being measures using geographically aggregated objective and subjective measures gleaned from mobile data in the United Kingdom. (Lai et al., 2019) review relevant work aiming at measuring human mobility and health risk in travelers using mobile geo-positioning data. (Bauer & Lukowicz, 2012) describe initial results from an ongoing project to use mobile phone sensors to detect stress related situations. (Canzian & Musolesi, 2015) seek to answer whether mobile phones can be used to unobtrusively monitor individuals affected by depressive mood disorders by analyzing only their mobility patterns from GPS traces. These early findings motivate a deeper evaluation of the contribution of population activity and mobility features to health disparity in cities.

Another limitation of the existing literature is the lack of consideration of non-linear interactions among features that could modulate the health status of urban neighborhoods. The existing spatial statistics methods assume a linear relationship between features, such as public parks. (Liu et al., 2017). In fact, urban health status is an emergent property arising as a result of the interactions among heterogeneous neighborhood features. However, the existing spatial and statistical approaches are unable to capture non-linear relationships among urban features that contribute to urban health gradients. This limitation could be addressed by spatial graph deep-learning techniques which could capture heterogeneous features of neighborhoods as well as their spatial interactions. Existing approaches do not provide a quantitative way to examine the similarities and discrepancies in the determinants of health disparities across different cities. The ability to juxtapose the determinants of health disparity across different cities could inform the transferability of urban design and planning strategies across different cities.

Recognizing these gaps, in this study, we examine features related to the built environment, population activities and mobility, and socio-demographics in training and testing graph attention network (GAT) models which could predict four major preventable threats to public health: obesity, diabetes, cancer, and heart disease. Population activity and human mobility data are collected from commercial location intelligence providers (Spectus and SafeGraph) and contain anonymized and aggregated data related to population visitations to points of interest (POI) as well as micro-mobility characteristics (such as average distance traveled and radius of gyration). In addition to population activity and mobility, we considered built environment features related to the density of facilities in neighborhoods, as well as air pollution and socio-demographic features. The GAT models treat the census tracts as nodes and edges, a construct which represents the spatial adjacency of the census tracts. The models were trained and tested in five U.S. cities across the four disease types. The models trained on a particular city were evaluated in other cities to evaluate the inter-city transferability of urban health determinants. Also, we conducted the GraphLIME method to rank the important features to specify the top five determinant features for each city and disease type. We also performed cross-city comparisons to reveal similarities and discrepancies in determinants of inter-city health disparity.

## Data collection and data processing

In this study, we focused on metropolitan counties in the United States. The selections were made based on population size, geographic distribution, and dataset availability. First, counties should have a population that is large enough to have intra-city health status variation. Second, the selected counties should be located in different regions in the United States to capture geographic variations. Finally, we chose Cook County (Chicago metro) in Illinois, Wayne County (Detroit metro) in Michigan, Fulton County (Atlanta metro) in Georgia, Suffolk County (Boston metro) in Massachusetts and Queens County in New York. For capturing population activity and mobility features in these counties, we selected the period of February 2020 which represents a steady-state period with no events that could influence population activity and mobility. Since human mobility data for the period before 2019 is either not available or very sparse and, because patterns of population activity and mobility is stable in cities and do not change from year to year, we used the February 2020 data for specifying the features.

**Public health data** *Obesity, Diabetes, Cancer, Heart Disease*

Public health data were collected from CDC (the Centers for Disease Control and Prevention). The dataset contains all United States at county, place, census tract, and ZIP Code Tabulation Area (ZCTA) levels. The dataset includes 29 health measures: 4 chronic disease health risk behaviors, 13 health outcomes, 3 health status, and 9 on preventive services. This study uses the data at the census tract (CT) level and used the four most common health conditions as dependent variables: obesity, diabetes, cancer, and heart disease. All health conditions were captured for adults aged older than 18 years in each census tract: obesity rate, percentage of diagnosed diabetes, percentage of cancer (excluding skin cancer), and percentage of coronary heart disease. The latest public health data released in 2021 is based upon 2019 Behavior Risk Factor Surveillance System (BRFSS) data(*Behavioral Risk Factor Surveillance System*). Although mobility features below use the data from February 2020, the lag in time would not influence the analysis since urban mobility patterns do not change significantly from year to year. We separated each health outcome into four classes uniformly into 25% quantile, 50% quantile, and 75% quantile. In this way, we can define every census tract in four levels of prevalence in terms of health statuses from 1 to 4, with one being lowest disease prevalence.

Fig. 1 illustrates an overview of the feature groups considered in this study. We extracted these health status features for each city at the census tract (CT) level. In addition to population activity and mobility features as inputs, we also considered features related to the built environment, environmental air pollution, and socio-demographics as reported in the literature. In particular, we considered the density of types of POIs in census tracts, as the density of POIs has been shown to have effects on health status. For example, (Galiatsatos et al., 2018) reported the relationship between social-economic status, tobacco store density, and health outcomes at the neighborhood level in a large urban community, and (Mason et al., 2018) shows strong associations between high densities of physical activity facilities and lower obesity for adults in mid-life. (Hasthanasombat & Mascolo, 2019) discussed the effect of the number of sports facilities on health status. (Horn et al., 2021) revealed the relationship between the fast-food environment and health status. which gives us some ideas to explore some other common facilities that could affect health status. Unlike these papers just find the effect of the neighborhood areas. Based on the existing literature, we define various features obtained from different datasets as inputs to our models (Table 1).

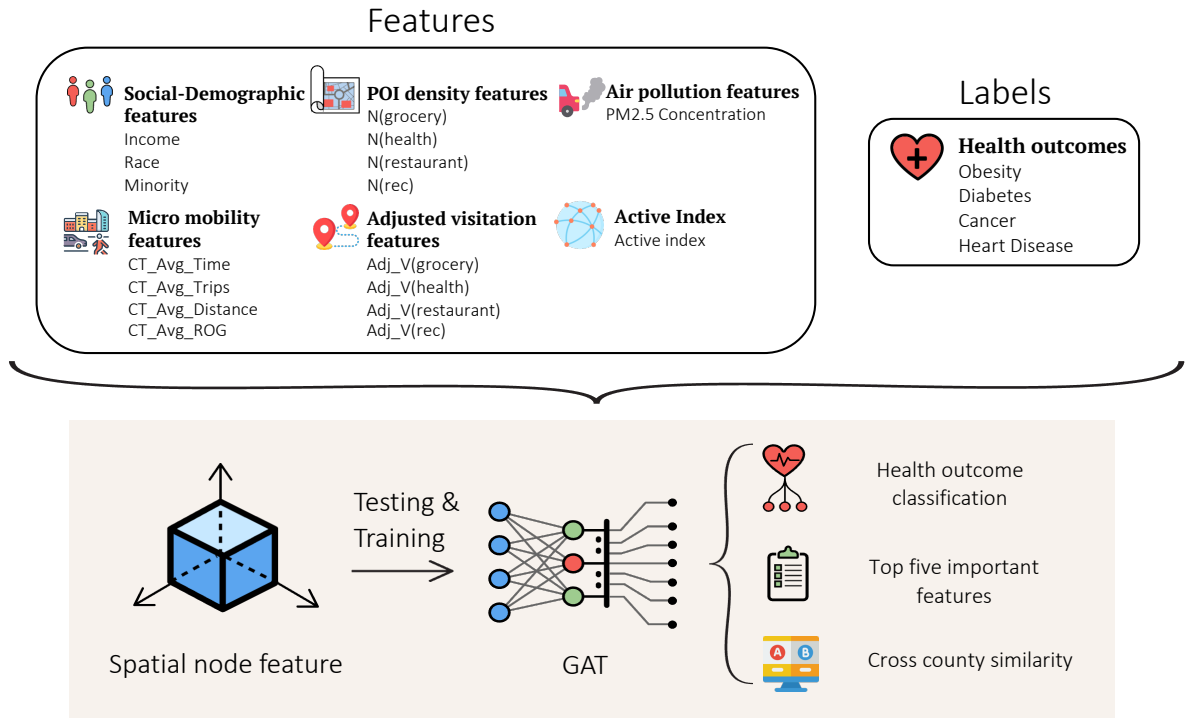

*Figure 1. Illustration of the analytical framework. The framework comprises two components: (1) feature groups and labels, and (2) training and testing process and three other analyses. The upper part of the figure is a schematic overview of feature groups and labels. The features and labels were collected at the census tract level. Each health status is labeled by the extent level. The lower part of the figure shows the training and testing process and three other analyses. (1) Using the trained GAT model to classify the extent of four health outcomes among five counties. (2) We used GraphLIME to identify the top five important determinants through input features and predicted classification results. (3) Analyze the model similarity across different counties by applying the original model and the transfer model.*

**Socio-Demographic** *Age, Income, Minority*

The socio-demographic data were retrieved from American Community Survey database administrated by US Census Bureau("United States Census Bureau,"). Age represents the percentage of people older than 65. Income stands for household income. Minority captures the percentage of minorities; Here minority groups refer to African Americans, American Indians, Alaska Natives, Asians, Native Hawaiian, and other Pacific Islanders.

**POI density** N(grocery), N(health), N(restaurant), N(rec), N(sport)

We quantified the number of different POI facilities in each CT using the SafeGraph POI data. In this study, we selected four POI categories: grocery facilities, health facilities, restaurants, and recreation centers. SafeGraph's core dataset contains spatial coordinates and addresses of POIs, as well as essential data about every POI, such as brand, and NAICS code industrial and commercial classifications. SafeGraph records codes and official full titles for the areas, subsectors, enterprises. Sorting by the prepared NAICS code table (Table SI-1 in the supplemental information), we can obtain the number of POI facilities in census tracts (CT).

**Air pollution** *Emission*

We used PM 2.5 emission data, a measure of particulate matter, released by Centers for Disease Control and Prevention (CDC). The dataset provides modeled predictions of PM 2.5 levels. It contains an estimated mean predicted concentration at the census tract (CT) level.

**Micro-mobility** *CT_Avg_Time, CT_Avg_Trips, CT_Avg_Distance, CT_Avg_ROG*

All measurements of micro-mobility are made at the level of individual residents of a census tract. The *CT_Avg_Time feature* captures the average travel time. The *CT_Avg_Trips* feature calculates the average number of trips. *CT_Avg_Distance* captures the average travel distance per census tract. We also used *CT_Avg_ROG (radius of gyration)* to capture the mobility extent for census tracts. The extent of human mobility can be captured based on the radius of gyration (ROG). *CT_Avg_ROG* can be calculated by the following equation,

$$rog = \sqrt{\frac{1}{N}\sum_{i=1}^{N}(p_i - p_{centroid})^2} \qquad \text{Equation (1)}$$

$$p_{centroid} = \frac{1}{M}\sum_{i=1}^{M} p_i \qquad \text{Equation (2)}$$

where $p_i$ is the *i*th position for the specific person, *p_centroid* is the central position of the person according to his moving trajectories. *M* is the number of trips completed by a user. *N* is the number of users in this census tract.

**Visitation features** *adj_V(grocery), adj_V(health), adj_V(restaurant), adj_V(rec)*

In addition to human mobility features, the number of visitations to each facility could be associated with health status. We normalized the visitations by dividing the populations by the number of persons in a census tract to adjust for the varied population size in census tracts. The adjusted visitations to different types of POI facilities can then be used as another set of population activity features.

The visitation data is collected primarily from the dataset furnished by location intelligence company Spectus. Spectus collects data from roughly 15 million daily active cell phone users in the United States. Compared to traditional call detail record (CDR) data, those tagged with highly accurate GPS information can pinpoint traveler routes and destinations. Accurate GPS information is extremely useful for detecting detailed visitations. Spectus' main database is built with data collected by third-party apps that capture cell phone location points with user consent. In a single day, third-party apps record more than 100 data points from a single person. The collected coordinates are attached to each POI in the database as a tuple of longitude and latitude. The location data from Spectus is de-identified to protect privacy.

To calculate human visitation to POI facilities, we had to identify each anonymous user's home at the census-tract level. Human home data were retrieved from the device table in the Spectus core database. A human's home will be recognized if the records show the stay time is more than 12 hours. Then we applied Spectus's stop table to specify the visitation pattern for each anonymous device_id. Stop points are captured in the dataset if individuals stay at a location for a while. Accordingly, we obtain the stop's location coordinate, stop-by date, and time for each stop point. Because the Spectus dataset lacks NAICS code information to distinguish the type of each POI, we complemented the dataset with the SafeGraph data. As mentioned above, SafeGraph data includes NAICS code and polygon information for each POI. The polygon information from Spectus and the NAICS information can be merged using the location coordinates, the tuple of longitude and the latitude, as the reference key. Therefore, each visited POI is tagged with a unique NAICS code. Then, for users living in each census tract, we aggregated the weekly visits to each POI.

**Active Index**

The active index is calculated using the following equation:

$$AI_i = \frac{In\_degree_i + Out\_degree_i}{Population\_size_i}$$  Equation (3)

Here, *i* is the *i*th census tract. *In_degree* is the number of users who visit this census tract. *Out_degree* is the number of users who leave this census tract. The active index is calculated by the normalized human flows. A larger active index indicates that the census tract is more active.

*Table 1. Summary of feature groups and notations in the model.*

| Socio-Demographic | POI density | Air condition | Micro-mobility | Visitation | Active Index |
|---|---|---|---|---|---|
| Age | N(grocery) | Emission | CT_Avg_Time | Adj_V(grocery) | Active_index |
| Income | N(health) | | CT_Avg_Trips | Adj_V(health) | |
| Minority | N(restaurant) | | CT_Avg_Distance | Adj_V(restaurant) | |
| | N(rec) | | CT_Avg_ROG | Adj_V(rec) | |

## Method

### GAT

The graph attention network model(Veličković et al., 2017) is a novel approach to processing graph-structured data by neural networks, leverages attention over a node's neighborhood. GAT is shown to achieve state-of-the-art results on various networked data, such as transudative citation network tasks and an inductive protein-protein interaction task. In this study, GATs utilize the spatial information of the node directly during the learning process. The first step performed by the graph attention layer is to apply a linear transformation-Weighted matrix **W** to the feature vectors of the nodes. To improve the stability of the learning process, multi-head attention is employed. We computed multiple attention maps and finally aggregated all the learned representations on one node using formula (5), where K is the number of independent attention maps used. Attention coefficient $\alpha$ determines the relative importance of neighboring features to each other.

$$e_{ij} = a(\mathbf{W}h_i, \mathbf{W}h_j) \quad \text{Equation (4)}$$

Due to the complicated connections between nodes, each node may have a different number of neighbors. To keep the same scale across all neighbors, the attention coefficients need to be normalized and then activated by the LeakyReLU function, where $N_i$ is the number of neighborhoods of node i.

$$\alpha_{i,j} = softmax(e_{ij}) = \frac{exp\,(e_{ij})}{\sum_{k \in N_i} exp\,(e_{ik})} \quad \text{Equation (5)}$$

$$\alpha_{i,j} = \frac{exp\,(LeakyReLU(a^T[\mathbf{W}h_i][\mathbf{W}h_j]))}{\sum_{k \in N_i} exp\,(LeakyReLU(a^T[\mathbf{W}h_i][\mathbf{W}h_k])} \quad \text{Equation (6)}$$

In this way, the new node's features can be represented by:

$$\vec{h_i}' = \sigma(\sum_{j \in N_i} \alpha_{i,j} \mathbf{W} \vec{h_j}') \quad \text{Equation (7)}$$

To improve the stability of the learning process, multi-head attention is employed. We computed multiple different attention maps and finally aggregated all the learned representations on one node, where K is the number of independent attention maps used:

$$\vec{h_i}' = \sigma(\frac{1}{K}\sum_{k=1}^{K}\sum_{j \in N_i} \alpha_{i,j}^k \mathbf{W}^k \vec{h_j}') \quad \text{Equation (8)}$$

**GraphLIME**

An important step in adopting deep learning models for urban design and planning purposes is to ensure that models are sufficiently explainable to inform plans and decisions about the importance of different features. In this study, we adopt GraphLIME for specifying the features' importance in the models. (Huang et al., 2022) proposed GraphLIME, a local interpretable model for graphs using the Hilbert-Schmidt Independence Criterion (HSIC) Lasso (Climente-Gonzalez et al., 2019), which is a nonlinear feature selection method. It takes into consideration numerous nonlinear aggregations. In the case of node classification, this embedding is used to separate nodes into classes. The core concept behind GraphLIME is to give a target node in the input graph and the node's N-hop vicinity to gather characteristics. Here, *N* is the number of layers in the trained graph neural network. Based on the weights of features in HSIC Lasso, GraphLIME can select important features to explain the HSIC Lasso predictions. Those selected features are regarded as explanations of the original GAT prediction. We focus primarily on the top five features and examine whether population activity and mobility features are among the most important.

**Cross-county similarity**

For the purpose of recognizing the similarities of the determinants of health status across different cities, we trained a GAT model using data from one county, (county A), then we saved this model's parameters along with the top five important features when it reached a good test accuracy. We call this a transfer model for all other counties other than county A. We then insert input features from another county into the transfer model, calculate the test accuracy and then compare the spatial structure of the transfer model against with original model. If the test accuracy is high and most of the areas are assigned to the same cluster between the two models, we can conclude that these two counties can share the same model and top five important features.

# Results

In this section, we present the results of multiple experiments to answer the following research questions: 1) to what extent population activity, human mobility, built environments, and air pollution features can predict the status of urban-scale health? 2) what is the importance of population activity and mobility features in predicting the health status of neighborhoods across the four diseases? 3) to what extent do models trained in one city could transfer to other cities to inform about the transferability of urban design and plans to promote urban health?

**Prediction performance**

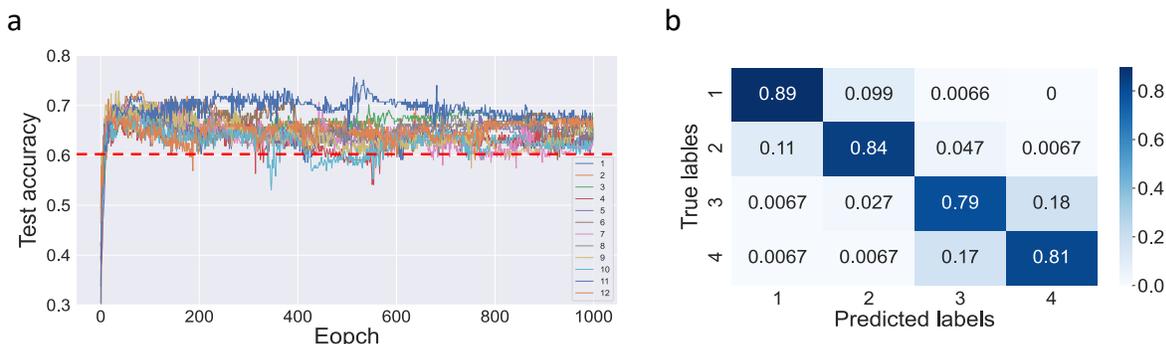

*Figure 2. Model performance for obesity in Wayne County. **a.** Results of test accuracy across 1000 epochs for 12 different hyperparameter sets. It indicates that the GAT model has a good performance by using defined feature groups to predict different health statuses. **b.** The confusion matrix. (The model used here is selected with the highest test accuracy according to a.)*

In this study, we used test accuracy, which is a metric for the predictability of the model. Taking the example of Wayne County, for each disease type, we classified the census tracts into four clusters by their quantiles. Accordingly, we gave each census tract a label indicating the extent of disease prevalence, with label 1 indicating the lowest percentage, and label 4, the highest. With the labels and input features, we trained the GAT model with all node features and complete adjacency matrix, but randomly chose 70% of the nodes as a training set for the supervised learning. Here, we set four hyperparameters in total: the number of hidden layers, the number of heads, the number of epochs, and the learning rate. The number of hidden layers and the number of heads can enrich the model capacity and stabilize the learning process. The learning rate controls how quickly the model adapts to the problem. We tuned these hyperparameters of the model: hidden layers (5, 10, 15), heads (3, 4, 5), and learning rate (0.025, 0.05). Figure 2 (a) shows the process of training and the improvement across the epochs for all nodes. The x-axis is the number of epochs; the y-axis is the test accuracy. We can see from the figure that models with different hyperparameter sets converged at different speeds. Test accuracy across epochs in different hyperparameter sets was greater than 0.6. Figure. 2b gives an example of classification results for the obesity rate in Wayne County by the confusion matrix. As shown in the figure, the GAT model has a good performance; clusters 1, cluster 3, and cluster 4 have an accuracy greater than 0.8. Even though the performance for cluster 3 is lacks the accuracy of the other clusters. The model can still predict more than half of the nodes precisely. By observing test accuracy in different hyperparameter sets, we chose the hyperparameter sets that show the highest test accuracy for our model. In this way, the extent of health status by using the defined feature groups can be quantified.

*Table 2. Model performance in different counties across four health statuses.*

|  | Obesity | Diabetes | Cancer | Heart Disease |
|---|---|---|---|---|
| **Cook, IL** | 0.737 | 0.737 | 0.679 | 0.625 |
| **Wayne, MI** | 0.757 | 0.724 | 0.613 | 0.586 |
| **Fulton, GA** | 0.689 | 0.754 | 0.672 | 0.656 |
| **Suffolk, MA** | 0.763 | 0.712 | 0.661 | 0.593 |
| **Queens, NY** | 0.706 | 0.619 | 0.619 | 0.536 |

Table 2 summarizes the model performance in counties across four health statuses. Even while some test accuracy is less than 0.6, the majority are greater than 0.6, which further demonstrates how successfully the input features combined with their non-linear relationships to predict the four main diseases at the census-tract level.

**Feature Analysis**

In the next step, we examined the importance of features in predicting the health status across different diseases to examine whether population activity and human mobility features are among the top features. We first trained the model by using each single feature group as input separately. We used the aforementioned hyperparameter set to train the model. The outcomes of predicting obesity across five counties are shown in Table 3. (The outcomes related to other health statuses can be found in the supplemental information (Tables SI-2, SI-3, and S-4). We found that the performance of the models using some single-feature groups, such as the POI density feature in Fulton County for obesity, even outperforms than using all features in predicting health status; however, the difference is not significant and the models with more features provide better explainability about features to inform urban design and planning to promote urban health. We also notice that in most cases among four diseases, models with socio-demographic features results in the best performance. This result confirms health disparity in urban areas. Evaluating disease prevalence based on sociodemographic data, however, provides limited insights for urban design and planning to promote urban health. The differences between the best feature groups in predicting health statuses motivated us to explore the combination of features in different feature groups as input features, but to focus on more than a single feature group, so that the models and their feature importance can inform urban design and planning to promote urban health.

*Table 3. Model performance in different counties for obesity by different feature groups.*

| County | Socio-demographic | POI density | Air condition | Micro-mobility | Visitation | Active Index | All-features |
|---|---|---|---|---|---|---|---|
| Cook, IL | 0.711 | 0.689 | 0.504 | 0.653 | 0.653 | 0.499 | 0.737 |
| Wayne, MI | 0.702 | 0.691 | 0.591 | 0.586 | 0.702 | 0.613 | 0.757 |
| Fulton, GA | 0.639 | 0.705 | 0.557 | 0.672 | 0.705 | 0.623 | 0.689 |
| Suffolk, MA | 0.644 | 0.746 | 0.644 | 0.678 | 0.712 | 0.661 | 0.763 |
| Queens, NY | 0.670 | 0.675 | 0.397 | 0.583 | 0.619 | 0.485 | 0.706 |

In the next step, we used GraphLIME as a method to provide a deep insight into feature importance in the prediction results. Generally speaking, each node (census tract) gives us a list of ranked features according to its contribution weight. Aggregating the contribution weight by all nodes in one county gives the overall feature importance. Figure 3 (a), (b), (c), (d) show examples of feature importance for Wayne County regarding obesity. The x-axis is the contribution weight for prediction; the y-axis shows the name of the features. Each box includes 601 aggregated nodes, which is the number of census tracts in Wayne County. Figure 3 depicts the ranked feature importance for obesity. We can see the top five features are the percentage of people older than 65, the percentage of minorities, the total income, radius of gyration, and the number of restaurants in census tracts. This result shows that some population activity and mobility features are among the most important factors alongside socio-demographic features.

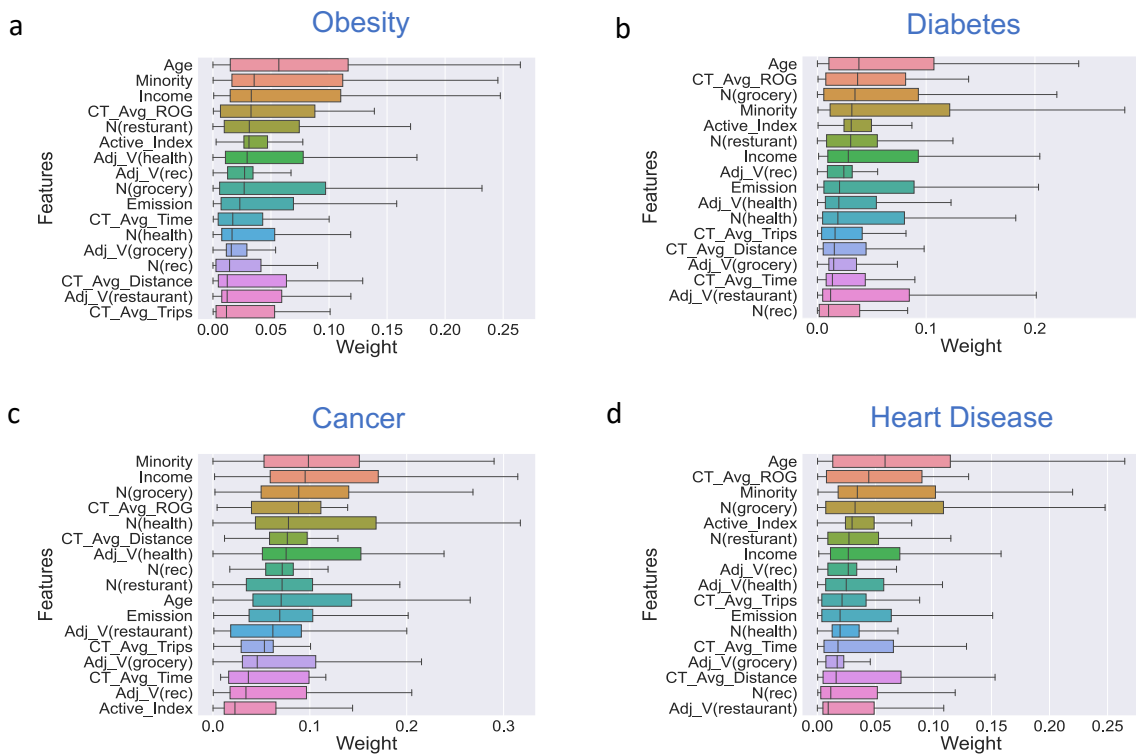

*Figure 3.* Feature importance in different disease types in Wayne County: *(a)* obesity, *(b)* diabetes, *(c)* cancer, and *(d)* heart disease. Each box shows the median and variation of contribution weight. The importance of features in each health status is sorted from largest to smallest.

*Table 4. Model performance in different counties across four diseases by top five important features.*

| Features groups | Feature | 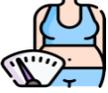 | 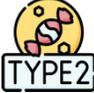 | 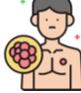 | 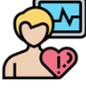 |
|---|---|---|---|---|---|
| Socio-demographic | Age | C W S | C W S Q | S Q | C W S Q |
|  | Income | W F S Q | C F S Q | C W S Q | C F S Q |
|  | Minority | C W F S Q | C W F S Q | C W F S Q | C W F S Q |
| POI density | N(grocery) | S Q | S W Q | C W S Q | W Q |
|  | N(health) |  |  | W |  |
|  | N(restaurant) | W |  |  |  |
|  | N(rec) | F Q | Q | F Q | Q |
| Air condition | Emission | C F Q | F | C F | C F S |
| Micro-mobility | CT_Avg_Time |  | C |  |  |
|  | CT_Avg_Trips | C |  |  |  |
|  | CT_Avg_Distance |  |  |  |  |
|  | CT_Avg_ROG | W | W | W | W |
| Visitation | Adj_V(grocery) |  |  |  | F |
|  | Adj_V(health) |  | C F | F | C |
|  | Adj_V(restaurant) |  |  |  |  |
|  | Adj_V(rec) |  | S |  | S |
| Active index | Active index | C F S | W F | C F S | W F |

*C: Cook County, IL. W: Wayne County, MI. S: Suffolk County, MA. F: Fulton County, GA. Q: Queens County, NY

By picking the top five important features to predict health status, test accuracy is very close to using all features as input. We can conclude that the top five features are enough to depict health status (see supplemental information, Table SI-5). By seeing Table 4, for obesity, we can observe that the minority feature is an important feature in all counties. This indicates that there exists a very strong relationship between the percentage of minorities and obesity, which is an evidence of health inequality for racial minorities in terms of obesity. Total income appears as an important feature in models of four counties, percentage of emissions, and active index appear in models of three counties. When we add the emission feature to the other top five features for each model, the model prediction improves. This result indicates that there is a strong relationship between emission and other top five features in predicting health status. Except for these features, the number of grocery stores and the number of recreation centers (such as gyms) appear twice in the models. CBG_Avg_Trips and CBG_Avg_ROG just show up once among the top five important features across all models. These results show the importance of socio-demographic features and the built environment features (number of certain POIs) in predicting obesity.

For diabetes, the percentage of minorities is an important feature in all five counties. The percentage of people older than 65 and total income appears four times in the top five features. Visits to health facilities, number of grocery stores, and active index appear twice. Visits to recreational centers, the number of recreational centers, the number of restaurants, emission, CBG_Avg_Time appears only once. Based on the result, minority, income, age, visits to health facilities, number of grocery stores, and active index are the most important features in predicting diabetes prevalence. CBG_Avg_Time, emission, the number of grocery stores, the number of restaurants, and visits to recreation centers are not major features in predicting diabetes. Compared with obesity, except for the socio-demographic features, visits to health facilities become more important, and emissions become less important. This result is intuitive as areas with more prevalence of diabetes would have a greater number of visits to health facilities since diabetes is known to exacerbate other health conditions.

For cancer, the number of minorities is the most important feature in all five counties. Total income and number of grocery stores appear four times. The active index shows up three times. Emissions, the number of recreational centers, and the percentage of people older than 65 appear twice. Visits to restaurants, visits to health facilities, and CBG_Avg_ROG only appear once. Based on these results, minority, income, number of grocery stores, active index, age, emissions, number of grocery stores, and number of recreation centers are the important features in predicting cancer. Different from the other disease types, the density of POIs is shown to be an important determinant in the prevalence of cancer across urban areas.

For heart disease, the percentage of minorities is still the most important feature. The total income and the percentage of people older than 65 appear four times. Emissions appears three times. Visits to grocery stores and active index show up twice. Visits to health facilities, recreation centers, and restaurants, the number of recreation centers, CBG_Avg_ROG just appears once. Based on the result, the percentage of minorities, the total income, age, emission, visits to grocery stores, and active index are important features in predicting heart disease. Health research has shown the significance of diet for heart disease. Visits to grocery stores could capture aspects of people's diet patterns. It is noteworthy to find visits to grocery stores among the top five important features for heart disease prediction.

In conclusion, the interactions among minority, age, income, POI density and POI visitations provide reliable predictions for all four disease types. The visitation features are more important in predicting diabetes and heart disease. For cancer, except for socio-demographic features, POI density is an important determinant for improving prediction performance. In addition, we also found that compared with other disease types, the prediction accuracy of obesity is the highest, and the distribution of important features is very concentrated, basically gathered in socio-demographic and POI density. For heart disease, the test accuracy of the model is the lowest, and the distribution of important features is very scattered, basically involving all feature groups.

**Cross-county similarity of determinants among health status**

Till now, we have illustrated that we can use socio-demographic population activity, human mobility, built environments, and air pollution features to predict urban-scale health status. Also, we have found the top five significant features for each health status for different counties. What if just one model were used, and the chosen top five features developed in one city applied to multiple cities? It is apparent that health status in different cities has distinct determinants. Nevertheless, this might be different in high-dimensional space, which can capture a more complex structure among the determinants of health status. Identifying the similarities of the determinants of health status is useful for examining the transferability of urban design and planning strategies that promote urban health across different cities. As shown above, the GAT model can capture complex non-linear relationships in health status between cities, which prompts us to explore the similarity of the top five determinants in different health status among different cities by implementing the original model and the transfer model.

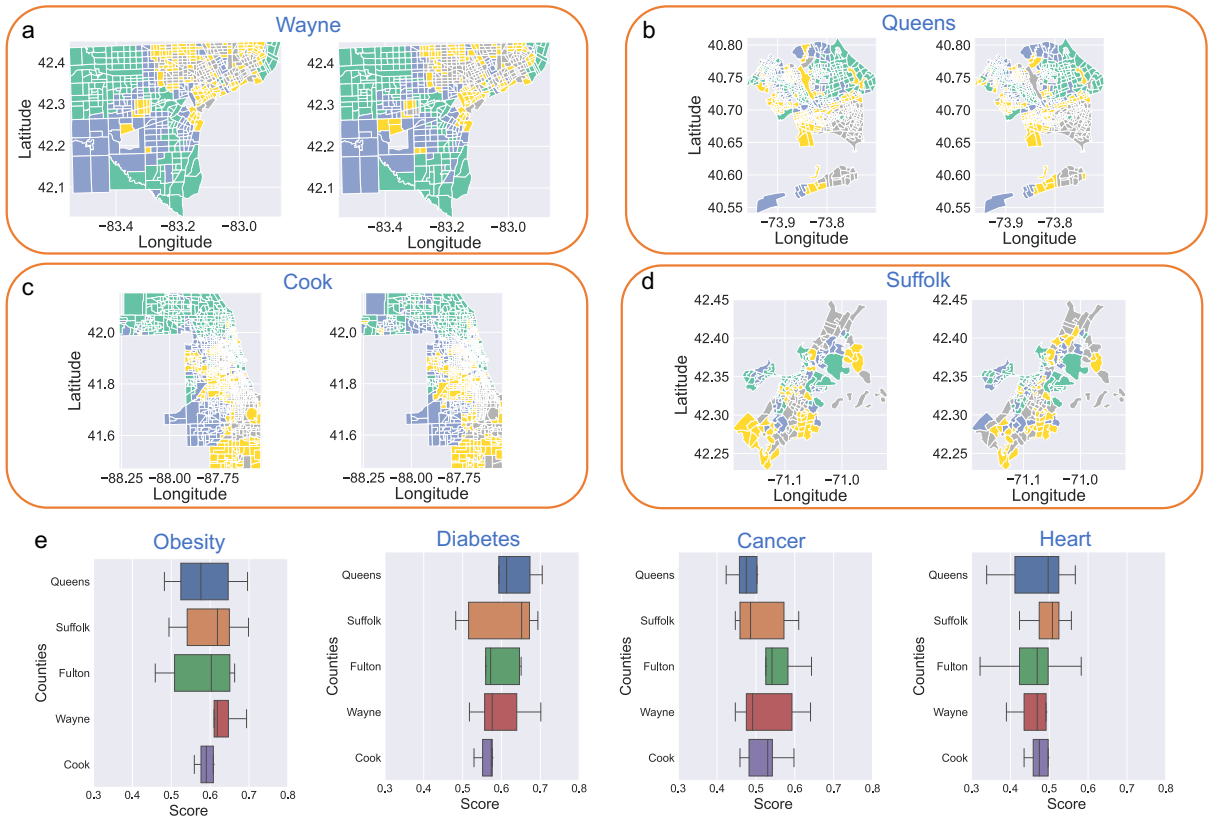

*Figure 4.* *(a) Geographical visualization of the classification results by implementing the Wayne County model as the original model and Queens County model as the transfer model for obesity. (b) Geographical visualization of the classification results by implementing the Queens County model as the original model and Suffolk model as the transfer model in Wayne County for Diabetes (c) Geographical visualization of the classification results by implementing the Cook County model as the original model and Queens county model as the transfer model in Cook County for Cancer (d) Geographical visualization of the classification results by implementing Suffolk County model as the original model and Queens County model as the transfer model in Queens County for obesity. (e) Test accuracy predicted by transfer model in different health statuses, from left to right: obesity, diabetes, cancer, heart disease. Each box shows the median and variation of test accuracy using the model trained by the transfer model for the other five counties. The results indicate that the transfer model and determinants can predict health status in high accuracy.*

First, we compared the spatial distributions of the clusters by using the original model and the transfer model. Figure 4 depicts some examples of spatial distribution of obesity by implementing the original model and the transfer model for different disease types. For instance, as shown in Figure 4 (a), the original model was trained in Wayne County, and the transfer model was trained in Queens County. We can find that the majority of the pairs of areas are assigned in the same clusters for Wayne County. The lower rate of obese people is clustered in the south and northwest parts, while the higher rate of obesity is in the northeast part of Wayne County. Similarly, the results shown in Figure 4 (b) indicate that most pairs of areas are assigned in the same clusters. However, in Figure 4 (c), we can find some different spatial distributions in the south and the northwest part of Cook County. In this case, the Queens County model does not share similar health status determinants with Cook County according to the spatial distribution plot. The same results can be observed in Figure 4 (d). Since the geographical structures differ significantly by using the original model and transfer model, it would be impractical to apply the same policies from Queens County to Cook County. These findings also encourage us to determine the degree to which the geographical structures of other counties are comparable.

Next, we explored quantifying the degree of similarity among cities in terms of different health statuses. As mentioned in the Method section, we first trained a GAT model in one county, for example, Cook County, until the model performance reaches the criteria requirements. This GAT model will be defined as a transfer model and later applied to the other four counties. We use the test accuracy of the transfer model to measure the extent of cross-county determinants similarity. In this way, we have four values in each box, each value representing the transfer model performance in the other four counties. Figure 4(e)–(h) provides strong evidence that the cross-county model can capture the complex non-linear relationships among the determinant features in predicting health status with more than 50% test accuracy. In other words, cities have 50% to 70% similarity in terms of the top five feature determinants predicting obesity and diabetes rate, 45% to 60% similarity for cancer, and 40% to 55% similarity for heart disease. Among all disease types, obesity shares the most similarity in terms of the determinants across the five-county model, and heart disease has the least similarity. At the county level, Suffolk County shares a large number of similarities with other counties regarding obesity, diabetes, and heart disease. Fulton County shares a number of similarities with other counties for cancer. These results may be influenced by the selection of counties in our study and may change with the addition of more counties. In conclusion, the results show the capability of the proposed method to explore cross-county similarities in examining the determinants of urban health disparity in high dimensional space. Besides, the transfer model could provide a comparison regarding the extent of similarities and differences in the determinants of health status among different cities.

## Discussion

In this study, we presented a graph deep learning approach for unveiling the determinants of urban health disparities. The findings of this study show that the combination of socio-demographic, POI density, and population activity features and their interactions are the main determinants in predicting the prevalence of obesity, diabetes, cancer, and heart disease. The results also show population activity features (point of interest visitation) are more important in predicting diabetes and heart disease, and POI density is an important feature for cancer prediction. Finally, the results related to the assessment of cross-city transferability show that, to a great extent, the core determinants of four disease types are similar; a model trained in one city can predict the spatial structure of disease prevalence in other cities.

This study presented in this paper provides important advancements in the way urban health disparity is examined. First, departing from the existing studies, which focus primarily on socio-demographic and environmental exposure features, our model captures heterogeneous features related to population activity and mobility, as well as POI facility density to specify the extent to which these urban characteristics contribute to the prediction of major disease types. Second, the GAT model provides a novel approach for examining urban health disparities as an emergent property arising from interaction among various heterogeneous urban features. Third, the approach used in this study provides a quantitative way to compare urban health disparity and the similarity of important determinants across different cities. Such quantitative approach is essential for evaluating the transferability of urban design and planning strategies, as well as public health policies across different cities. Through these advancements, the model and results presented in this study enable more integrated urban design strategies to promote health equity in cities. For example, the results of this study could inform urban design policies related to facility distribution and facility visitation as they are shown to be among the important determinants of disease prevalence. Broadly, this study contributes to the growing field of urban artificial intelligence (urban AI) for integrated urban design.

Some limitations within this study, which would need to be overcome in the future. First, additional urban features could be considered in the models. In this study, we focused on features related to population activity and POI facility density, since there was a dearth of studies examining these features and their contribution to urban health status. Future studies can expand the type and number of features to enhance the model presented in this paper. Second, the explainability of graph deep learning models is still a technical challenge to be solved. We used the state-of-the-art method of GraphLIME to specify feature importance. However, GraphLIME cannot specify the important feature interactions, nor the sign of features. With further advancements in the field of explainable graph deep learning, future studies could further examine the sign of determinants of urban health status, and also specify the important feature interactions to inform policies.

The most relevant limitation is that without knowing the sign of each feature, GraphLIME can calculate important weight but cannot provide the feature sign. It would be very challenging to explain what the feature sign and its relation to the health status. Future research should resolve this problem. And during the process of model training, we found that performance would be improved if the urban area had more nodes, in other words, more census tracts. Future research should focus on collecting data with a larger sample of census tracts.

## Author contributions

C.L and A.M designed research; A.M and C.L collected data; C.L analyzed data; C.L, A.M and C.F wrote and revised the paper.

## Data availability

The data that support the findings of this study are available from SafeGraph and Spectus, but restrictions apply to the availability of these data, which were used under license for the current study. The data can be accessed upon request submitted to the data provides. Other data we use in this study are all publicly available.

## Code availability

The code that supports the findings of this study is available from the corresponding author upon request.